\documentclass[10pt,twocolumn,letterpaper]{article}

\usepackage[pagenumbers]{wacv} 

\newcommand{\shortname}{GenHSI\xspace}
\newcommand{\coolname}{\shortname}

\usepackage[most]{tcolorbox}

\newtcolorbox{promptbox}{
  colback=blue!5,      
  colframe=blue!50!black, 
  boxrule=0.5pt,       
  arc=4mm,             
  fontupper=\sffamily, 
  left=6pt,right=6pt,top=6pt,bottom=6pt
}

\definecolor{wacvblue}{rgb}{0.21,0.49,0.74}
\usepackage[pagebackref,breaklinks,colorlinks,allcolors=wacvblue]{hyperref}

\def\wacvPaperID{2386} 
\def\confName{WACV}
\def\confYear{2026}

\title{\shortname: Controllable Generation of Human-Scene Interaction Videos}

\author{Zekun Li
\quad
Rui Zhou
\quad
Rahul Sajnani
\quad
Xiaoyan Cong
\quad
Daniel Ritchie
\quad
Srinath Sridhar
\\
Brown University
}

\begin{document}
\newcommand{\teaserCaption}{
}
\twocolumn[{
    \renewcommand\twocolumn[1][]{#1}
    \maketitle
    \centering
    \begin{minipage}{1\textwidth}
        \centering 
        
        \includegraphics[width=\linewidth]{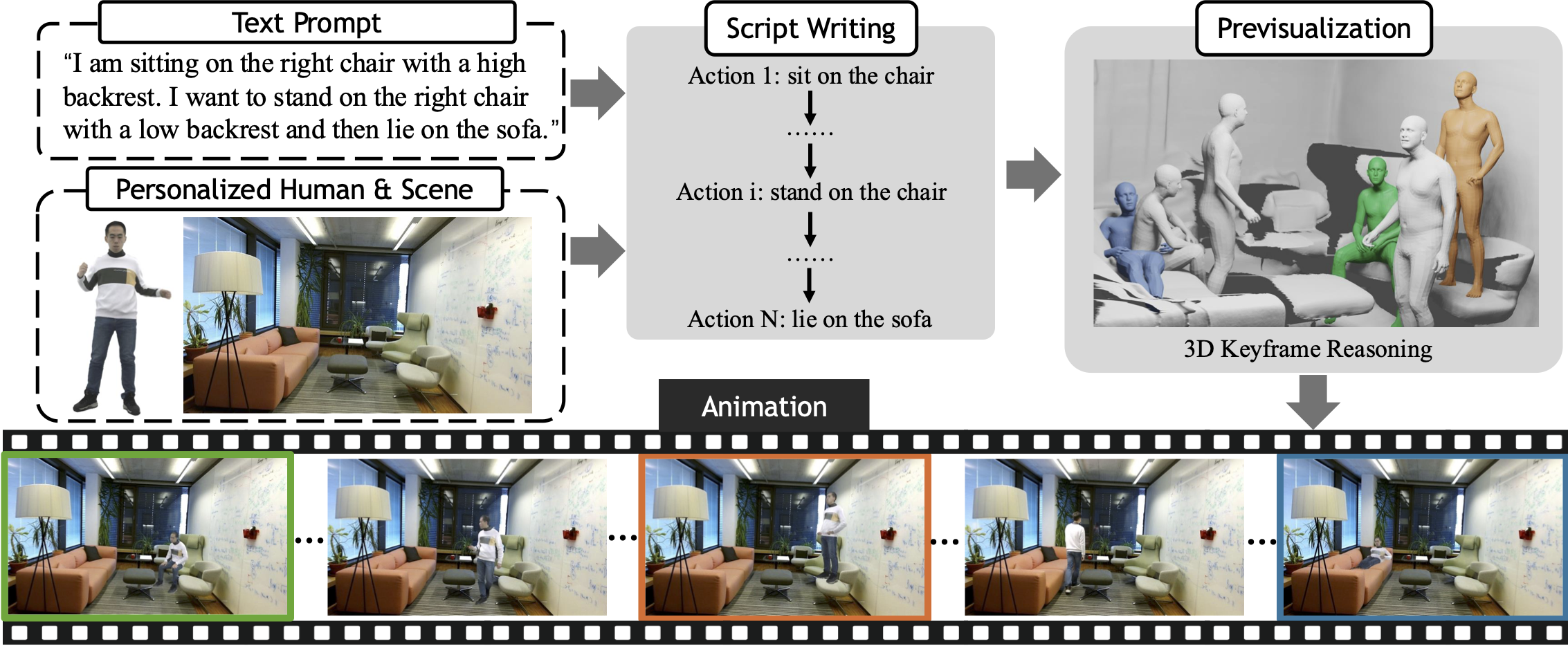}
    \end{minipage}
    
    \captionsetup{type=figure}
    \vspace{-2mm}
    \captionof{figure}{\teaserCaption \textbf{\coolname{}} is a 3D-aware controllable human-scene interaction (HSI) video generation method.
    We mimic the real-world filmmaking procedure, \ie, Script Writing, Previsualization, and Animation, to generate an extendable HSI video clip with arbitrary lengths of action chains.
    Given images of the scene and character with the action sequence prompt, our method will render multiple 3D-aware keyframes based on the posed 3D Gaussian avatar and 3D Gaussian scene.
    Finally, we interpolate them into a continuous video using the pretrained video diffusion model.
    The frames with colored borders are selected 3D-aware keyframes that map to the color human meshes.
    }
    \label{fig:teaser}
    \vspace{5mm}
}]
\begin{abstract}
Large-scale pre-trained video diffusion models have exhibited remarkable capabilities in diverse video generation.
However, existing solutions face several challenges in generating long videos with rich human–scene interactions (HSI), including unrealistic dynamics and affordance, lack of subject identity preservation, and the need for expensive training.
To this end, we propose \shortname, a training-free method for controllable generation of long HSI videos with 3D awareness.
Taking inspiration from movie animation, we subdivide the video synthesis into three stages: (1) script writing, (2) pre-visualization, and (3) animation.
Given an image of a scene and a character with a user description, we use these three stages to generate long videos that preserve human identity and provide rich and plausible HSI.
Script writing converts a complex text prompt involving a chain of HSI into simple atomic actions that are used in the pre-visualization stage to generate 3D keyframes.
To synthesize plausible human interaction poses in 3D keyframes, we utilize pre-trained 2D inpainting diffusion models to generate plausible 2D human interactions based on view canonicalization, which eliminates the need for multi-view fitting in previous works. 
We then extend these interactions to 3D using robust iterative optimization, informed by contact cues and reasoning from VLMs.
Prompted by these 3D keyframes, the pretrained video diffusion models can better generate consistent long videos with plausible dynamics and affordance in a 3D-aware manner.
We are the first to synthesize a long video sequence with a chain of HSI actions without training based on the image references of the scene and character.
Experiments demonstrate that our method can generate HSI videos that effectively preserve scene content and character identity with plausible human-scene interaction from a single image scene. 
\end{abstract}    
\section{Introduction}
Rapid progress has been made in the last few years on the problem of photorealistic image~\cite{rombach2022high,esser2024scaling,blackforestlabs_flux,Chen2023PixArtFT} and video~\cite{hacohen2024ltx, blattmann2023stable, ma2025step, Guo2023AnimateDiffAY, Chen2024VideoCrafter2OD, Lin2024OpenSoraPO, Wang2023LAVIEHV, Xing2023DynamiCrafterAO} generation, especially using diffusion models~\cite{ho2020denoising,song2020denoising}.
These generative image and video models can synthesize high-quality and diverse content and even support multi-modal control using text instructions~\cite{ge2023preserve,ho2022video,blattmann2023align}, audio~\cite{sung2023sound}, camera poses~\cite{he2025cameractrl, wang2024motionctrl}, and other modalities~\cite{zhang2023adding,guo2024i2v,ma2024cinemo,xing2024dynamicrafter, Shi2024MotionI2VCA, chen2024foundhand}.
These developments have opened up broad applications:
for instance, in personalization~\cite{ruiz2023dreambooth,chen2024dreamcinema}, style transfer~\cite{Chung_2024_CVPR}, and editing~\cite{sajnani2024geodiffuser,shi2024dragdiffusion,zheng2024oscillation}.
%
Particularly popular are applications that enable pose-controllable generation of humans in videos~\cite{Feng2023DreaMovingAH, Cai2023GenerativeRC}, human-object interactions~\cite{Peng2023HOIDiffTS, Zhu2024DreamHOISG,min2024genheld, Xu2024AnchorCrafterAC}, and 3D human motion generation~\cite{Li2024ZeroHSIZ4}. 

Despite this progress, current video diffusion models (VDMs) face several challenges, particularly when dealing with human physical interactions with the environment.
Since the dynamics are implicitly formalized in visual generative models, they often result in unrealistic physical phenomena~\cite{motamed2025generative}, especially when multifaceted dynamics happen, \eg unrealistic HSI.
Additionally, without explicit 3D modeling, VDMs frequently fail to ground text instructions in accurate spatial reasoning, resulting in artifacts such as characters heading in incorrect directions or hallucinated 3D affordances during HSI.
Furthermore, it is hard to generate videos with a consistent person identity when composing image references of the human and scene implicitly, even with expensive training or fine-tuning~\cite{chen2025multi, Kim2025DAViDMD}.

To this end, we present \textbf{\shortname}, a training-free method for controllable generation of long human-scene interaction (HSI) videos via 3D-aware keyframe prompting given an image of the scene and person with the HSI text prompt (see \cref{fig:teaser}).
Specifically, we break down the HSI video generation problem into three stages: (1)~script writing, (2)~pre-visualization, and (3)~animation.
In the \textbf{script writing stage}, \shortname prompts a VLM~\cite{openai2025chatgpt4o} to generate a more detailed \emph{script} with step-by-step instructions that decompose complex interactions into a sequence of simpler atomic tasks.
In the \textbf{pre-visualization stage}, we achieve accurate human-object contacts and interactions by creating 3D keyframes for each atomic task.
This novel keyframe generation step synthesizes human-object interactions using an inpainting model in canonical view and optimizes contacts between them in 3D for improved affordance.
A key advantage of our work is the benefit we get from large reconstruction models~\cite{xiang2024structured, Wang2024MoGeUA} (even when they provide inaccurate geometries) to reconstruct 3D scenes from real-world images, alleviating the need for accurate scene geometry assumed in prior works~\cite{li2024genzi, cong2024laserhuman,zhao2022compositional,jiang2024scaling,wang2024move}.
Finally, during the \textbf{animation stage}, the scene~\cite{Kerbl20233DGS} and character~\cite{qiu2025lhm} are modeled as 3D Gaussians, which inherently accommodate visual occlusion in 3D.
We interpolate the rendering results based on parsed HSI scripts to generate controlled 3D-aware videos using an off-the-shelf video generation model~\cite{klingai-frames}.
As there are no existing solutions that generate long videos with accurate human-object interactions, we compare individual components of our system against prior works.
Our evaluation includes comparing human-object interaction estimation against diffusion-based solutions~\cite{parihar2024text2place} and 3D human-scene interaction methods~\cite{zhao2022compositional,wang2024move,li2024genzi} on real-world scanned scenes~\cite{hassan2019resolving}. 
Additionally, we also compare against commercial solutions~\cite{klingai-elements}.
Extensive qualitative and quantitative experiments demonstrate that our method produces more physically plausible results in 3D human-scene interactions and achieves superior visual quality in preserving human-scene consistency.

In summary, our main contributions are summarized as:
\begin{itemize}
    \item We propose a training-free method \textbf{\shortname} for controllable generation of long human-scene interaction videos given only a scene image, a person image, and an interaction text description.
    \item Rather than deal with the HSI video generation using a single stage, we break it down into three stages: \textbf{script writing}, \textbf{pre-visualization}, and \textbf{animation} that enable better 3D control through 3D-aware keyframe prompting, while being training-free and retaining person identities.
\end{itemize}
These contributions open the doors to diverse applications in controllable long HSI video generation, training data generation, and video personalization.



\section{Related Works}
\paragraph{Human-Scene Interaction Video Generation:}
Current video models can generate impressive human motions~\cite{Fang2024MotionCharacterIA} with personalized visual characters~\cite{Zhang2023PIAYP, Liu2025PhantomSV, Zhang2025MagicMI}, with controls such as human poses~\cite{Gan2025HumanDiTPD, Xu2024AnchorCrafterAC}, masks~\cite{Zhou2024MotionCF}, and audio~\cite{Lin2025OmniHuman1RT}. 
However, it is difficult to control the \emph{interaction} of the person with the environment.
Some existing models~\cite{chen2025multi,gu2025diffusion} still struggle to generate HSI with plausible affordances and dynamics, even when trained on large datasets with extensive GPU resources for fine-tuning.
To reduce the hallucination in HSI video generation, some approaches~\cite{Men_2025_CVPR,hu2025animate} leverage the existing video as the reference and replace either the human or the object to achieve customization.
However, these approaches rely on real videos as a foundation and are thus limited to editing existing content, lacking the ability to generate human-scene interactions from scratch.
Recent work~\cite{kim2025target} distills the interaction prior into a special token~\cite{ruiz2023dreambooth} to achieve customized HSI video generation from image references of scene and character, which still involves training and is limited to a small scale of video generation model.
Different from \shortname, our 3D-aware keyframe prompting is training-free and model-agnostic general solution for plausible HSI video generation. 

\begin{figure*}
    \centering
    \vspace{-0.3cm}
    \includegraphics[width=1\linewidth]{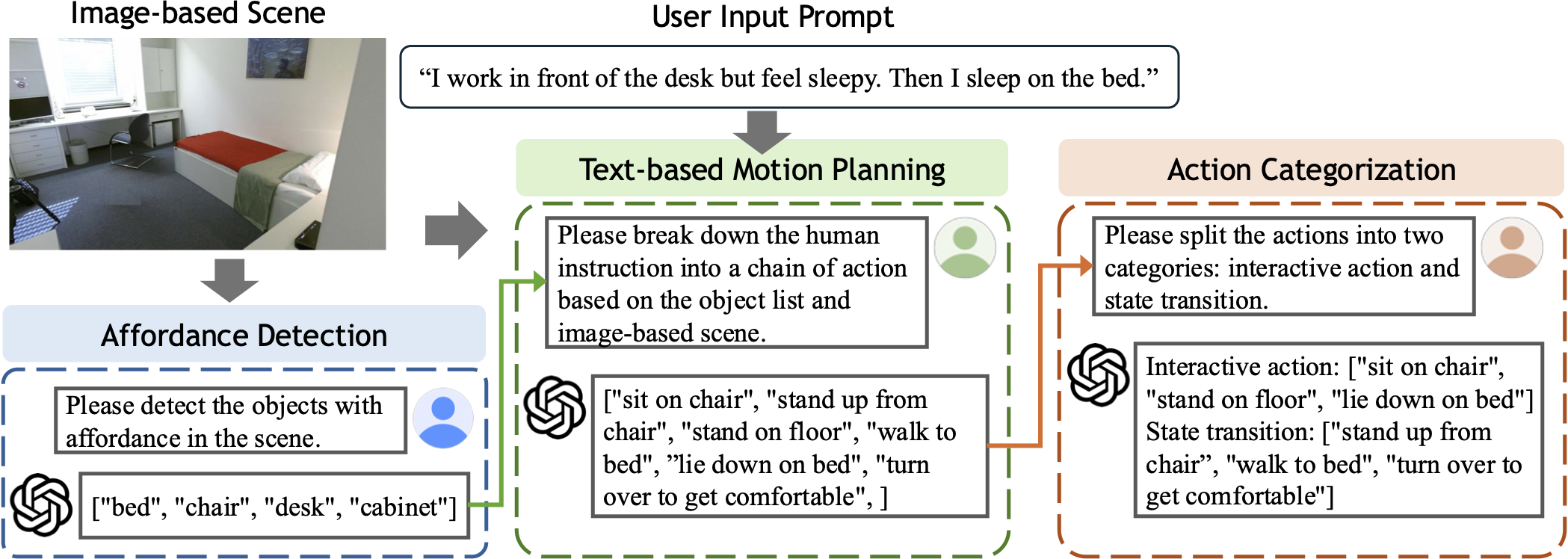}
    \vspace{-0.6cm}
    \caption{
    \textbf{Script Writing Stage:} Complex high-level text descriptions from users do not provide a detailed scene and task understanding for the desired long video generation.
    The script writing stage first \textbf{identifies and segments objects} that the human can interact with in the scene. 
    These objects, along with the given human prompt, are used to \textbf{perform text-based motion planning} from a VLM~\cite{openai2025chatgpt4o} that provides us with \textbf{interactive actions \& state transitions for keyframing} in the \textbf{Pre-Visualization Stage}.
    }
    \label{fig:motion-parsing}
\end{figure*}
\paragraph{Human-Scene Interaction 3D Pose Synthesis:}
Synthesizing humans in scenes is a crucial yet challenging task in computer vision and graphics, requiring the modeling of complex, high-level semantic understanding, such as affordances and interactions.
With paired scene-motion datasets~\cite{araujo2023circle,guzov2021human,hassan2019resolving,cong2024laserhuman,jiang2024scaling}, previous works~\cite{wang2024move,jiang2024scaling,zhao2022compositional,yi2024generating,cen2024generating,cong2024laserhuman,wang2022humanise,huang2023diffusion} encode the scene geometry as latent conditions for human pose and motion generation.
However, these methods rely on high-quality 3D scene datasets with motion-captured human interactions, making them difficult to scale and generalize across diverse environments.
In contrast, we focus on zero-shot interaction synthesis to generate plausible human-scene interaction from an image diffusion prior without any motion dataset or training, inspired by the comprehensive understanding of human-scene composition priors in diffusion-based image editing systems~\cite{ruiz2024magic,tewel2024add,parihar2024text2place,kulal2023putting,Yang_2024_CVPR}.
Although existing works~\cite{li2024genzi,kim2024beyond} attempt to lift 3D human poses from multi-view inpainting, they fail to guarantee cross-view consistency because each view is inpainted independently.
This 3D inconsistency often leads to instability in subsequent pose lifting optimization. 
To address this issue, we canonicalize coarse 3D reconstructions of objects and perform character inpainting from the canonical viewpoint that maximizes the visibility of object affordances. 
Our efficient 3D human lifting approach, grounded in chain-of-contact reasoning and single-view inpainting, effectively avoids these limitations.
\section{\shortname}
%

The goal of \shortname is to generate a personalized long human-scene interaction video given natural language text descriptions and an image of a scene and a character.
To overcome hallucinations during interaction in existing video generative models, we utilize 3D-aware keyframes generated from explicit 3D reasoning to prompt off-the-shelf video generative models.
Our solution mimics the real-world filmmaking process through a modularized approach (Script Writing - Previsualization - Animation) as shown in~\cref{fig:teaser}.
We first parse a high-level text description into simple atomic tasks using a VLM under image scene context (see \cref{fig:motion-parsing}), which includes physical interactions and state transitions based on the affordances in the image scene (\cref{sec:screenwriter}).
After obtaining the detailed and structured motion script, we prompt human-object interaction using a novel zero-shot pose generation that leverages a pre-trained inpainting model (\cref{sec:inpainting-stage2}).
We then compose the scene~\cite{xiang2024structured, Wang2024MoGeUA} and human~\cite{SMPL:2015,SMPL-X:2019} in 3D under affordance constraints that lead to physically plausible interactions between the character and the environment (\cref{sec:pose_opt}).
To render them into a holistic 3D-aware keyframes and interpolate as HSI video, we use the estimated depth point cloud~\cite{Wang2024MoGeUA} as initialization for scene 3D Gaussian fitting and feed-forward 3D Gaussian avatar generation model~\cite{qiu2025lhm} to obtain character assets (\cref{sec:animation}).

\begin{figure*}
    \centering
    \vspace{-0.2cm}
    \includegraphics[width=1\linewidth]{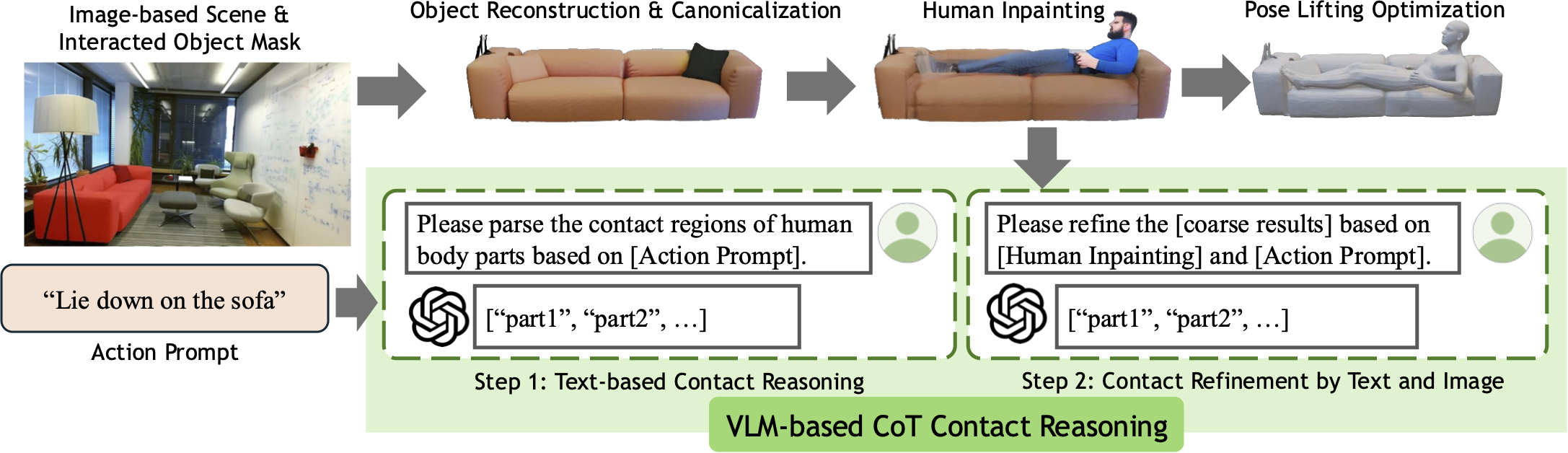}
    \vspace{-0.5cm}
    \caption{
    \textbf{3D Keyframe Generation for Pre-visualization.} \shortname synthesizes 3D human-scene interaction pose based on the pretrained 2D image inpainting diffusion model to create a 3D keyframe as an intermediate step for HSI video generation. 
    Our method lifts the 2D human inpainting result in the canonical view of the target object based on contact cues reasoned by the VLM chain-of-though. 
    }
    \label{fig:pose-generation}
    \vspace{-0.5cm}
\end{figure*}
\subsection{Script Writing Stage}
\label{sec:screenwriter}
The HSI text descriptions from users are usually structureless and need a chain-of-action to execute. 
Hence, we first perform script writing that takes a complex human description and a scene image to reason~\cite{openai2025chatgpt4o} and localize~\cite{wu2019detectron2,kirillov2023segment} the target interaction object using a segmentation mask.
Based on the image understanding, we break the text description into atomic action scripts based on the scene understanding in \cref{fig:motion-parsing}, which serves as the basis for generating 3D keyframes in the \textbf{Pre-visualization} stage and interpolating keyframes in the \textbf{Animation} stage.


\paragraph{Scene Understanding and Object Detection:}
Before generating the video, it is essential to understand the scene and identify objects that humans can interact with.
We first provide the input scene image to the VLM and prompt it to describe the scene as well as list objects that support human-object interaction in \cref{fig:motion-parsing}. 
According to the user input prompt, we perform open-set detection~\cite{liu2024grounding} \& segmentation~\cite{kirillov2023segment} to locate the involved objects.
The object masks $m_o$ will be used in 6D pose and scale estimation.


\paragraph{Text-based Motion Planning and Categorization:}
After obtaining the scene description and segments of possible objects of interest, our next step is to plan the progress of HSI in a plausible dynamics procedure.
This text-based motion plan is crucial for breaking down complex motion descriptions into simpler, detailed instructions that are used to create keyframes and maintain smooth motion in video generation.
Hence, we utilize the VLM to convert complex, high-level human text descriptions into a script of low-level, atomic actions that ensure smooth and natural transitions.
Each low-level action clearly describes the relationship between the human and the object instance in the scene using a verb or preposition. 
However, as shown in \cref{fig:motion-parsing}, not all the low-level actions from text-based motion planning outputs represent a physical interaction between the human and the scene, which motivates us to categorize them into actions that describe human-object interactions (interactive actions) and body movement without changing interaction (state transitions). 
As the interactive actions model physical constraints in human-scene interactions, we use them to create 3D keyframes to enhance the plausibility in video generation.
Additionally, utilizing state transitions, along with interactive actions, ensures a logical and smooth progression of human motion during keyframe interpolation.


\subsection{Pre-visualization Stage: 2D Human Inpainting}
\label{sec:inpainting-stage2}
\begin{figure}[ht]
    \vspace{-0.1cm}
    \centering
    \begin{subfigure}[b]{0.55\linewidth}
         \centering
         \includegraphics[width=\textwidth]{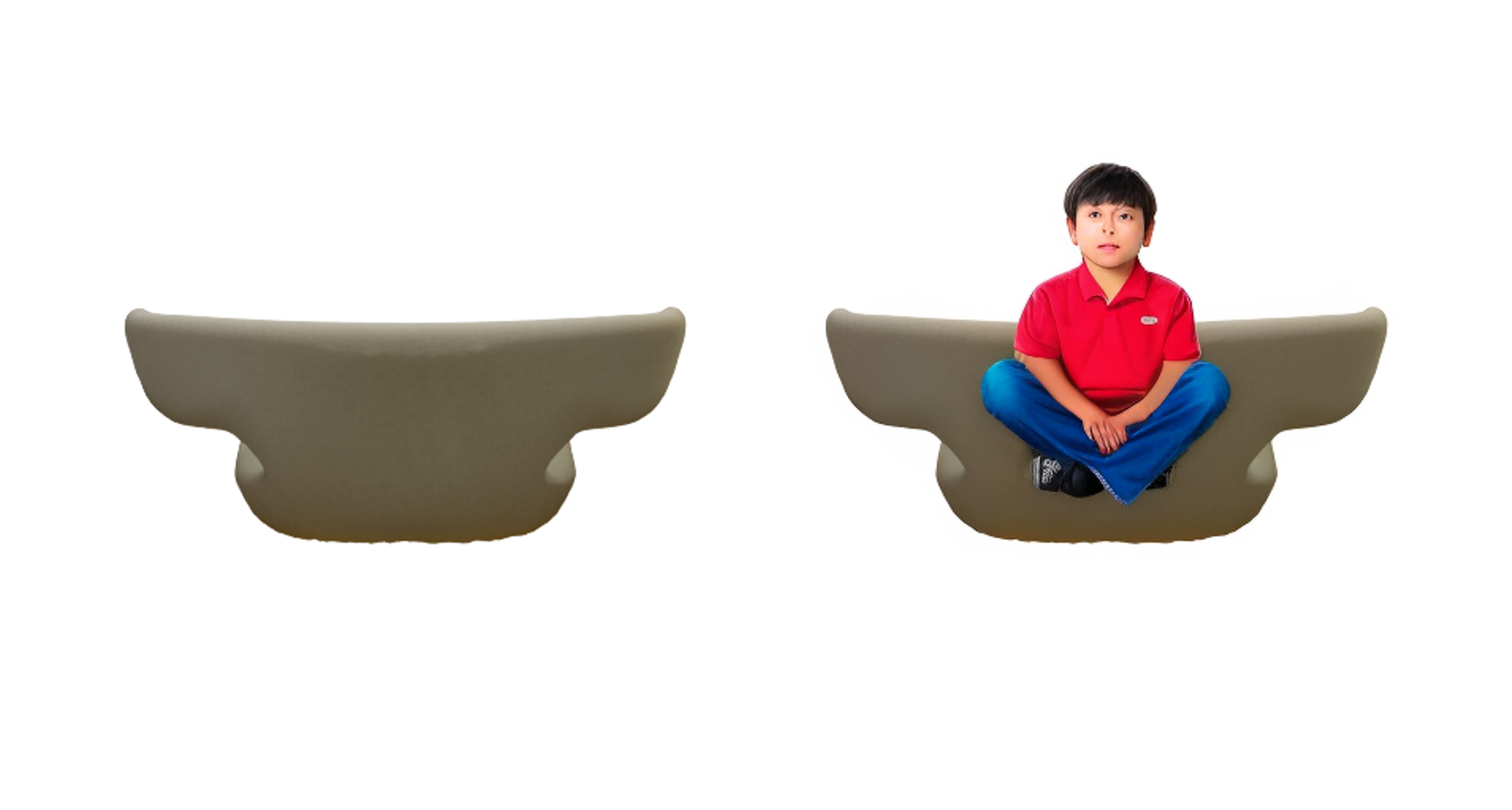}
         \vspace{-0.4cm}
         \caption{Hallucinate Affordance}
         \label{fig:fail1}
         \vspace{-0.3cm}
     \end{subfigure}
     \begin{subfigure}[b]{0.44\linewidth}
         \centering
         \includegraphics[width=\textwidth]{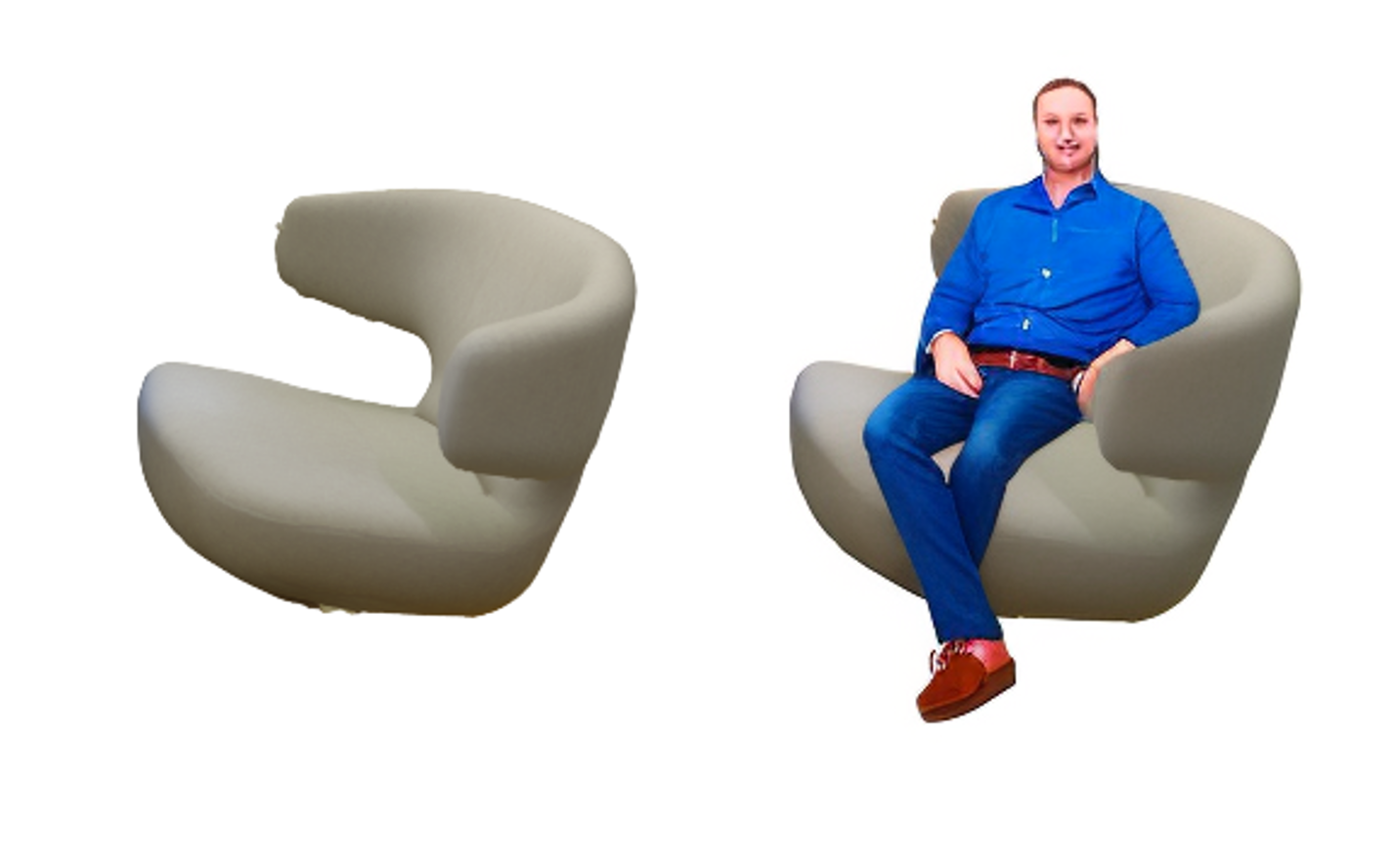}
         \vspace{-0.4cm}
         \caption{Inplausible Interaction}
         \vspace{-0.3cm}
         \label{fig:fail2}
     \end{subfigure}
    \caption{Failure cases when inpainting a human from different object views, demonstrating that the performance of pose generation from an image diffusion prior is view-dependent.}
    \vspace{-0.5cm}
    \label{fig:fail-case}
\end{figure}

In the second stage \textbf{Pre-visualization}, our method generates keyframes to prompt off-the-shelf video generative models.
To reduce hallucinations in spatial understanding during HSI video generation, such as 3D affordance and interaction targets, we propose a novel and efficient 3D HSI pose generation method using a pretrained 2D image inpainting model and contact-guided 3D pose lifting optimization.
As shown in \cref{fig:pose-generation}, given the image of the scene with the target object segment and the interactive action text prompt $c$ obtained from \textbf{Script Writing} stage, we synthesize a 3D human mesh $\mathcal{M}_h$ parameterized by pose and shape parameters $(\theta,\beta)$, performing the specified interaction with the object in the scene.
\paragraph{Variant Performance of 2D Human Inpaint:}
Different from regular inpainting tasks, inpainting humans in scene images requires localizing the human in the scene and prompting the model to interact with the human and the object of interest.
We input the object image and a prompt detailing the desired human-object contact to extract human-object contact priors from the diffusion model. 
However, directly obtaining human-object interaction in random views of objects does not always perform well (see \cref{fig:fail-case}).
This issue occurs primarily because: 1)~the diffusion model is biased towards forward-facing views of objects (\cref{fig:fail1}) and 2)~inpainting in random views often occludes human-object interactions, leading to sub-optimal human poses  (\cref{fig:fail2}).
For instance, inpainting hallucinates the human sitting in the air in \cref{fig:fail1} when the back of the chair is visible in the scene, and \cref{fig:fail2} displays a human sitting in an uncomfortable pose looking towards the camera, even when the couch is rotated sideways.
\begin{figure}[ht]
    \centering
    \includegraphics[width=0.9\linewidth]{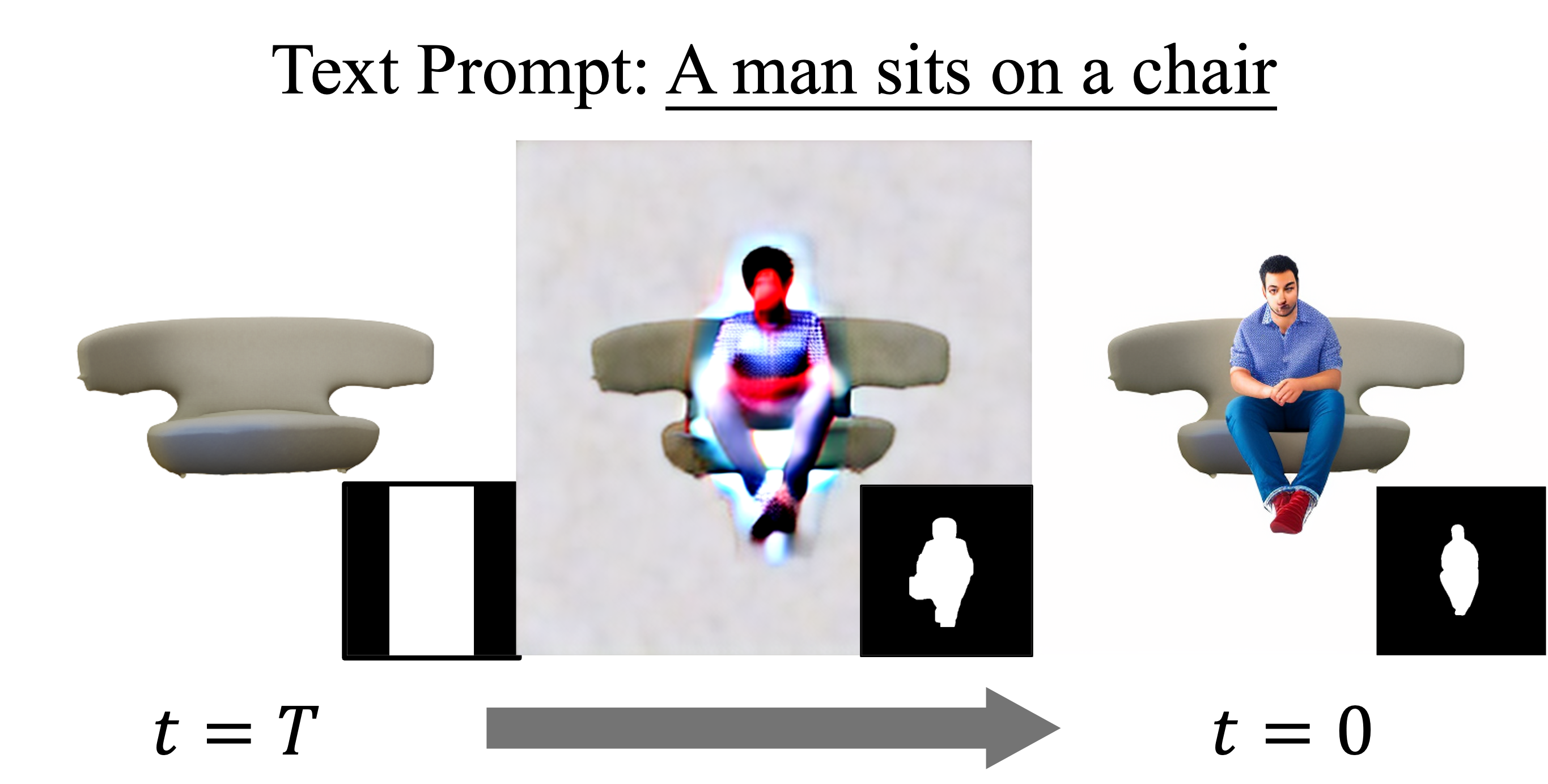}
    \vspace{-0.3cm}
    \caption{
    Human-Object Interaction from 2D Inpainting Models in the canonical view of objects. 
    We progressively update the human mask (bottom) while denoising the inpainting result (top). 
    }
    \label{fig:inpainting}
    \vspace{-0.4cm}
\end{figure}
\paragraph{Effective Human Inpainting in the Canonical View:}
Our solution to curb the afore-mentioned issues is to first canonicalize the object of interest and then synthesize human interaction poses.
We hypothesize that diffusion models can readily capture the canonical view of objects, making it more accurate to estimate human object interactions in this view.
Hence, we prompt inpainting models to synthesize human-object contacts in the canonical view of the object. 
We reconstruct the 3D object using the image-to-3D model Trellis~\cite{xiang2024structured} and canonicalize it via OrientAything~\cite{wang2024orient}. 
Next, we render the object to obtain the 2D human interaction in this canonical view using an inpainting model~\cite{inpainting-model} (see ~\cref{fig:pose-generation,fig:inpainting}) instead of rendering in random multiple views as previous works~\cite{li2024genzi,kim2024beyond}.
The canonicalized rendered image of the object is input to the diffusion model with a coarse mask $m_t$ to inpaint. 
We then predict the clean image at every denoising step using Tweedle's formula and use Detectron~\cite{wu2019detectron2} to estimate the human mask. 
This human mask is then used to update the input mask of the inpainting model for the next denoising step, leading to easy, efficient, and fine-grained interaction (see \cref{fig:inpainting}) as follows: 
\begin{align}
    \mathbf{z}_{0|t} &= \frac{\mathbf{z}_{t} - \sqrt{1 - \bar{\alpha}_t}\epsilon_\Theta(\mathbf{z}_{t}, \mathbf{z}_{0}^*, m_t, c, t)}{\sqrt{\bar{\alpha}_t}} \\
    m_{t-1} &= \text{Segment}(\text{Decode}(\hat{\mathbf{z}}_{0|t})) \\
    \hat{\mathbf{z}}_{0|t} &= (1-\downarrow\!m_{t-1})\odot \mathbf{z}_{0}^* + \downarrow\!\!m_{t-1} \odot \mathbf{z}_{0|t}\\
    \mathbf{z}_{t-1} &= \sqrt{\bar{\alpha}_t} \hat{\mathbf{z}}_{0|t} + \sqrt{1-\bar{\alpha}_t} \epsilon
\end{align}
where $\downarrow\!m$ is the downsampled mask with the aligned shape with noise latent $\mathbf{z}$ and $\mathbf{z}_0^*$ is the original object rendering image latent extracted by the U-net encoder in the pretrained diffusion model.
We then estimate the SMPL-X parameters by HybrIK-X~\cite{li2021hybrik,li2023hybrik} in the canonical view that offers rich visual affordances and reduced self-occlusions.



\subsection{Pre-visualization Stage: 3D Pose Lifting}
\label{sec:pose_opt}
Since we extract the human pose from a single-view image, there is depth-scale ambiguity after estimation.
To compose the human and target object seamlessly in 3D, we formulate an optimization framework that corrects the scale $s_h$, translation $t_h$, and global rotation $r_h$ of the human to resolve inaccurate scale \& depth in 3D.
This optimization (1) uses the contact cues from VLM to compose the human and object in 3D, (2) minimizes the penetration between the human and the coarse 3D object reconstruction, and (3) matches the silhouette of the projected human mesh with the inpainted 2D interaction result. 


\paragraph{Interaction Affordance Loss:}
To compose the human and the target interacting object in the same 3D space, we first use the VLM to reason which predefined body parts contact the target object, as shown in \cref{fig:pose-generation}.
Given the contact cues, we propose an interaction loss $\mathcal{L}_{hoi}$ to ensure that the regions highlighted by the VLM are in contact.
This loss is formulated by minimizing the loss between the points $P_{h}$ of contact on the human surface that are identified by the VLM and the points $P_o$ on the object surface.
We first gather the possible contact points in $P_{h}$ and $P_o$ using the k-nearest neighbor method to obtain $P_{h}^*$ and $ P_o^*$. 
We then ensure that the point pairs in $ P _ {h} $ and $ P_o$ exist in the k-nearest neighbor subsets of each other. 
This loss is formulated as
\begin{equation}
    \mathcal{L}_{hoi} = \sum_{x\in P_{h}^*} \min_{y \in P_o^*} \|x-y\|^2_2 + \sum_{y\in P_{o}^*} \min_{x \in P_h^*} \|y-x\|^2_2.
\end{equation}

\paragraph{Penetration Loss:} 
Interaction loss alone increases human object contact but leads to human-object penetration. 
Hence, we encourage the optimization to avoid human-object penetration by constructing a signed distance field (SDF) $\Phi$ from the human mesh $\mathcal{M}_h$ and ensuring that points $v$ on the object surface $\mathcal{M}_O$ have a non-negative SDF value \ie no penetration with the human mesh:
\begin{equation}
    \mathcal{L}_{pen} = -\mathbb{E}_{v\in \mathcal{M}_O}[\min\big(\Phi(v), 0\big)].
\end{equation}
\paragraph{Silhouette Loss:}
Both $\mathcal{L}_{hoi}$ and $\mathcal{L}_{pen}$ ensure better human-object affordances and contact, but they do not fix the scale of the human. 
We ensure that the projected mask of the human matches the inpainting mask using an intersection over union (IoU) constraint.
Our solution involves rendering two masks from a silhouette rasterizer $\Pi$ corresponding to 1) the human  $m_h = \Pi(\mathcal{M}_h)$ and 2) the human with object occlusions $m_{hoi} = \Pi(\mathcal{M}_h|\mathcal{M}_O)$.
We then ensure that these masks overlap with the inital human mesh projection $m_{h}^{init}$ obtained using SMPL-X and initial human inpainting mask $m_{hoi}^{*}$:
\begin{equation}
    \mathcal{L}_{mask} = \frac{m_h \cap m_h^{init}}{m_h \cup m_h^{init}} + \frac{m_{hoi} \cap m_{hoi}^{*}}{m_{hoi} \cup m_{hoi}^{*}}.
\end{equation}
Matching the mask without object occlusion alone results in improper projection, where the human mesh may translate away from the object to just match itself.
In contrast, matching only the mask with object occlusion does not provide information about the occluded regions.
Hence, we use both masks to penalize the optimization.
The optimization solves the scale $s_h$, translation $t_h$, and global rotation $r_h$ of the human mesh $\mathcal{M}_h$ for every keyframe to compose the human and target object in 3D space by minimizing a combined loss function:
\begin{equation}
    \mathcal{L}_{total} = \mathcal{L}_{hoi} + \mathcal{L}_{pen} + \mathcal{L}_{mask}.
\end{equation}


\subsection{Animation Stage}
\label{sec:animation}
In this stage, we render the 3D-aware keyframes to prompt the pretrained video generative model to generate plausible HSI videos based on the optimized 3D human pose and the parsed chain-of-action.

\paragraph{3D-aware Keyframe Rendering}
To insert the optimized 3d human pose $\{\theta, r_h,t_h,s_h\}$ into the camera space of the input image scene, we estimate the 6D pose and scale of the object according to the coarse 3D object reconstruction template.
The object 6D pose $\{r_o, t_o\}$ is first initialized by DINOv2 feature similarity~\cite{ornek2024foundpose}.. 
For better alignment between the 3D reconstructed object and the depth estimation of the image scene, we recalculate the scale $s_o$ and translation $t_o$ based on silhouette and depth alignment in the original image scene.
To insert the character in the human reference image with optimized pose into the image scene with a plausible visual effect, we represent both the human and scene in 3D Gaussians~\cite{Kerbl20233DGS}, which allows us to resolve occlusions in 3D space naturally. 
We initialize the 3D Gaussian centers for each pixel based on the monocular depth estimator~\cite{Wang2024MoGeUA} and optimize only the color, rotation, and scale parameters for each 3D Gaussian by rendering and comparing to the input image view of the scene.
The human 3D avatars are obtained via the existing feed-forward Gaussian avatar generation model~\cite{qiu2025lhm}.
Compared with using image editing models to add humans into the image scene~\cite{parihar2024text2place,tewel2024add}, leveraging explicit 3D representations allows us to render 3D-aware keyframes with minimal change to the background and high preservation of character identity, while naturally handling occlusions between humans and the environment, making them well-suited for prompting off-the-shelf video generative models.
\paragraph{3D-aware Keyframe Interpolation}
The Pre-visualization stage provides us with a 3D storyboard in the form of keyframes for our video.
These 3D keyframes provide us with rich human-object contacts for each interactive action, enabling the rendering of human-object interactions in the input view. 
After rendering the keyframes, we animate them using Kling AI 1.6~\cite{klingai-frames}, which generates transition frames between the start frame and the end frame using the state transition actions obtained from the VLM as described in ~\cref{sec:screenwriter}.
Interpolating keyframes is analogous to in-context prompting in language generation: by decomposing an HSI video into a sequence of salient keyframes, we introduce richer constraints that mitigate hallucinations.
Incorporating detailed atomic state transitions into keyframe interpolation helps maintain human identity consistency and enhances motion realism for HSI video generation.

\begin{figure*}
    \centering
    \vspace{-0.4cm}
\scalebox{0.9}{
    \includegraphics[width=1\linewidth]{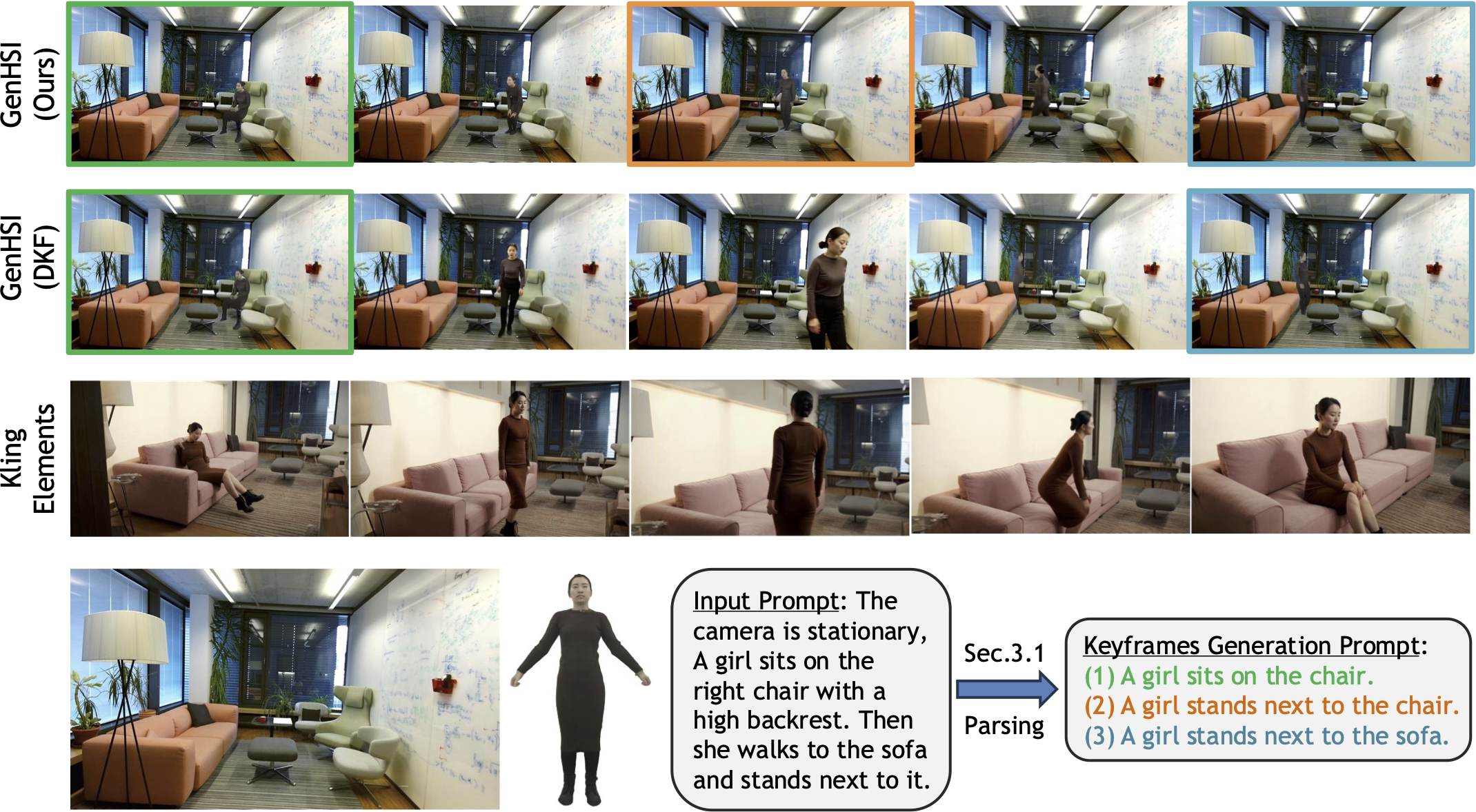}
    }
    
    \includegraphics[width=0.96\linewidth]{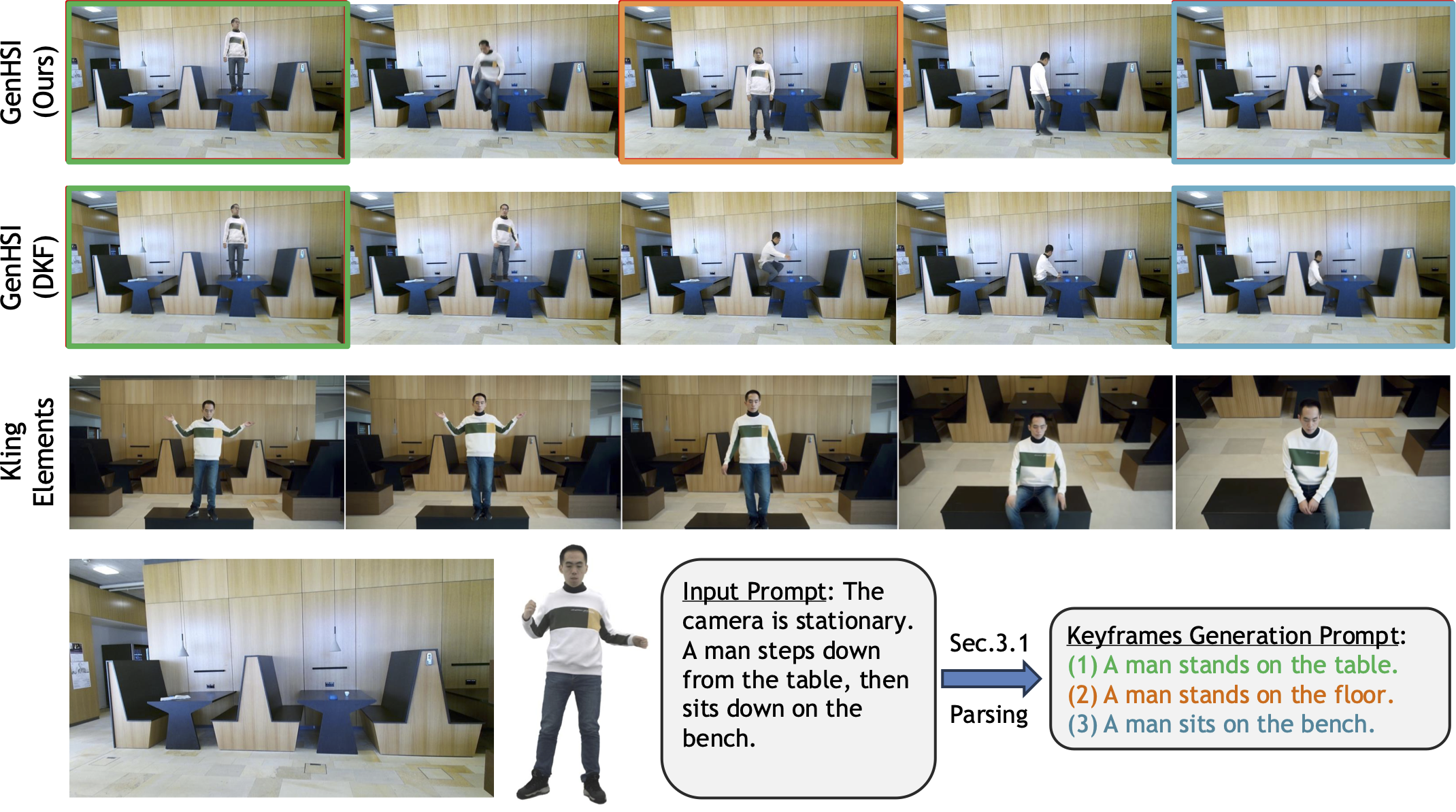}
    \vspace{-0.3cm}
    \caption{
    \textbf{HSI Video Generation Qualitative Results:} \shortname (ours) produces the best results with subject identity preservation and good human-object contacts.
    \shortname (DKF) only uses two keyframes (start and end frame) and often changes the identity of the subject.
    Kling Elements drastically change the scene and character. 
    Each video displays the keyframes highlighted with a red bounding box.
    }
    \label{fig:video_comp}
    \vspace{-0.4cm}
\end{figure*}
\section{Experiments}
Since no solutions exist that achieve the same goal of controlling long video generation with human-scene interactions, we instead conduct quantitative and qualitative evaluations to compare the components of \shortname with alternative solutions for 3D human pose generation and video synthesis involving human-scene interactions.

\paragraph{Dataset:}  
To fairly compare with previous pose generative methods~\cite{zhao2022compositional,li2024genzi,wang2024move}, we utilize a well-used 3D scene dataset PROX-S~\cite{zhao2022compositional,zhang2020generating} to demonstrate the effectiveness of our human pose generation.
Since our method uses the image as the representation of the scene, we filter out the invisible object interaction test cases in PROX-S based on the real-world images of the scenes for fair comparison.
To increase the scene diversity, we render the synthetic 3D scene dataset~\cite{li2024genzi,jiang2024scaling} and use Flux-depth to convert them into photorealistic image scenes. 
The composed HSI video generation evaluation set contains 50 samples.

\paragraph{Metrics:} We use community-accepted video evaluation metrics from~\cite{huang2023vbench}\footnote{We use the \href{https://github.com/Vchitect/VBench/tree/master/vbench2_beta_long}{long-VBench} to evaluate the HSI video since our generated video duration is larger than 10s.} to measure \textbf{Subject Consistency}, \textbf{Dynamic Degree}, \textbf{Background Consistency}, \textbf{Motion Smoothness}, and \textbf{Imaging Quality} in the long video results. 
\textbf{Subject} \& \textbf{Background Consistency} measure the semantic similarity of the subject and the background scene, respectively, using features extracted from DiNO. 
\textbf{Dynamic Degree} \& \textbf{Motion Smoothness} measure the average optical flow between consecutive frames within the video and their consistency across 
a video.
Additionally, we follow~\cite{zhao2022compositional, li2024genzi} and use community-accepted metrics of \textbf{Semantic Clip}, \textbf{Contact}, \textbf{Non-Collision}, \textbf{Entropy}, and \textbf{Cluster Size} to evaluate human-scene interaction quality.
We refer the reader to the papers referenced above for more details about each metric.

\subsection{Qualitative Results}
We demonstrate some HSI video generation results in different settings, shown as \cref{fig:video_comp}.
The frames with colored borders are the 3D-aware keyframes generated based on the atomic action prompt parsed in \cref{sec:screenwriter}.
In \shortname, we synthesize 3D-aware keyframes for each interactive action to prompt the off-the-shelf video generative model, achieving the least spatial hallucination and highest character identity preservation. 
To evaluate the effectiveness of 3D-aware keyframes prompting, we conduct an ablation setting with only dual keyframes (DKF), \ie the start frame and end frame.
Without detailed keyframe prompts to constrain the video generation, the generative model usually hallucinates the transition dynamics, \eg in the first sample (top), the woman goes out of the screen first and then magically reappears in the scene; in the second sample (bottom), the man is shown sitting down through an unrealistically narrow space between a table and a bench.
To demonstrate the advantage of \shortname, we compared it with a commercial solution, Kling-Elements~\cite{klingai-elements}, that supports composing multi-image references to generate customized videos. 
Although Kling-Elements can generally generate HSI video with better quality in terms of human appearance details, it cannot preserve the human identity like \shortname, \eg it changes the color of the cloth in the first test case.
Additionally, Kling-Elements seldom follows the text prompt to drive the human character to interact with the image scene.
Instead, it usually hallucinates the 3D scene and affordance, \eg the model creates a non-existent table and misinterprets it as the bench to sit on in the second test case.

\subsection{Quantitative Evaluation}
\paragraph{Inpainting Successful Rate in Different Viewpoints:}
To validate our assumption about 2D image inpainting models struggling with view-dependent performance in \cref{sec:inpainting-stage2}, we conduct a human judgment with 10 users across 30 inpainting results to evaluate the plausibility of inpainting in different views compared with the canonical view.
\begin{figure}[ht]
\vspace{-0.2cm}
\captionof{table}{ \textbf{Success Rate of Inpainting across Views}
\label{tab:sr_inpaint}
$\Delta \theta$ denotes the yaw angle relative to the canonical pose of the reconstructed 3D object.
}
\vspace{-0.11in}
    \centering
    \scalebox{0.7}{
\setlength{\tabcolsep}{6pt} 
\renewcommand{\arraystretch}{1} 
\begin{tabular}{rrrrr}
    \toprule
     
    & $\Delta \theta = 0$ & $|\Delta \theta| \in (0,\frac{\pi}{6}]$ & $|\Delta \theta| \in (\frac{\pi}{6}, \frac{\pi}{3}]$ & $|\Delta \theta| \in (\frac{\pi}{3}, \frac{\pi}{2}]$ \\
    \midrule
    Success Rate & 93.33\% & 54.67\% & 20\% & 6.67\%\\
    \bottomrule
\end{tabular}
}
\vspace{-0.7cm}
\end{figure}

\paragraph{3D Human-Scene Interaction:}
We evaluate the performance of our human-object interaction synthesis against GenZI~\cite{li2024genzi} \& COINS~\cite{zhao2022compositional} on filtered PROX-s, which exclude the invisible object interactions.
\cref{tab:3d_metrics} compares 3D Human Object Interaction results generated in the pre-visualization stage against baselines. 
\shortname improves Semantic Clip and Contact from the SOTA GenZI~\cite{li2024genzi} in zero-shot human scene interaction generation.
Benefiting from inpainting humans from the canonical view, the diffusion model efficiently inserts humans with plausible poses by only inpainting from a single view.
Our method can also perform comparably on other metrics to both~\cite{li2024genzi,zhao2022compositional} even when we do not have human joint pose optimization based on accurate 3D scenes.
Without canonicalization for human inpainting, the single-view human pose generation usually results in implausible HSI poses.

\begin{figure}[ht]
\captionof{table}{ \textbf{3D HSI Pose Generation}
\label{tab:3d_metrics}
``SV'' means only inpaint single view.
``MV'' means inpaint multiple views.
}
\vspace{-0.11in}
    \centering
    \scalebox{0.6}{
\setlength{\tabcolsep}{6pt} 
\renewcommand{\arraystretch}{1} 
\begin{tabular}{l|rrrrr}
    \toprule
     &\begin{tabular}[c]{@{}c@{}}\textbf{Semantic} \\ \textbf{Clip}\end{tabular} {\color{blue} $\uparrow$}  &  \textbf{Entropy} {\color{blue} $\uparrow$} & \begin{tabular}[c]{@{}c@{}}\textbf{Cluster}\\ \textbf{Size}\end{tabular}{\color{blue} $\uparrow$} & 
     \multicolumn{1}{c}{\begin{tabular}[c]{@{}c@{}}\textbf{Non-} \\ \textbf{Collision}\end{tabular}} {\color{blue} $\uparrow$} &
     \textbf{Contact}{\color{blue} $\uparrow$}\\
    \midrule
    \textbf{COINS (3D)}~\cite{zhao2022compositional} & \textbf{0.2624} & 2.695 & 0.813 & 0.974 & \underline{0.969} \\
    \textbf{GenZi (3D+MV)}~\cite{li2024genzi} & 0.2521 & \textbf{2.779} & \textbf{0.914} & \textbf{0.983} & 0.971\\
    \textbf{\shortname (SV)} & \underline{0.2578} & 2.601 & 0.852 & \underline{0.980} & \textbf{0.984}\\
    \textbf{\shortname (SV)} w/o can. & 0.2513 & \underline{2.734} & \underline{0.877} & 0.865 & 0.966\\
    \bottomrule
\end{tabular}
}
\vspace{-0.3cm}
\end{figure}

\begin{figure}
\captionof{table}{\textbf{Long-VBench Video Quality}~\cite{huang2023vbench} 
\label{tab:video_quality_metrics}
\shortname beats commercial video customization solution across major metrics, but shows lower Dynamic Degree as the consistent background does not contribute to the optical flow used in the evaluation. 
\shortname (DKF) - dual key frame improves over commercial model, but increasing keyframes in \shortname (Ours) improves consistency, subject identity, motion smoothness, and image quality.
}
\vspace{-0.05in}
    \centering
    \scalebox{0.5}{
\setlength{\tabcolsep}{6pt} 
\renewcommand{\arraystretch}{1} 
\begin{tabular}{l|rrrrr}
    \toprule
     &\begin{tabular}[c]{@{}c@{}}\textbf{Subject} \\ \textbf{Consistency}\end{tabular} {\color{blue} $\uparrow$}  & \multicolumn{1}{c}{\begin{tabular}[c]{@{}c@{}}\textbf{Background} \\ \textbf{Consistency}\end{tabular}} {\color{blue} $\uparrow$} & \begin{tabular}[c]{@{}c@{}}\textbf{Motion}\\ \textbf{Smoothness}\end{tabular}
     {\color{blue} $\uparrow$} & \begin{tabular}[c]{@{}c@{}}\textbf{Imaging}\\ \textbf{Quality}\end{tabular}{\color{blue} $\uparrow$} &      
     \begin{tabular}[c]{@{}c@{}}\textbf{Dynamic}\\ \textbf{Degree}\end{tabular}
     {\color{blue} $\uparrow$}\\
    \midrule
    \textbf{Kling AI 1.6 Elements}~\cite{klingai-elements} & 0.961 & 0.949 & 0.991 & 0.726 & \textbf{0.960}\\
    \textbf{\shortname (DKF)} & 0.972 & 0.960 & 0.993 & 0.740 & 0.885 \\
    \textbf{\shortname (Ours)} & \textbf{0.985} & \textbf{0.969} & \textbf{0.996} & \textbf{0.754} & 0.609 \\
    \bottomrule
\end{tabular}
}
\vspace{-0.5cm}
\end{figure}

\paragraph{Video Quality:} \shortname has the highest score in  \textbf{Subject Consistency} (\textbf{0.985}), \textbf{Background Consistency} (\textbf{0.969}), \textbf{Motion Smoothness} (\textbf{0.996}), and \textbf{Image Quality} (\textbf{0.754}) that beating commercial solutions.
Our videos exhibit limited \textbf{Dynamic Degree} due to their static backgrounds, which do not contribute to the optical flow used for dynamic degree measurement.
And we also evaluate the effectiveness of keyframing based on atomic interactive action parsing in \textbf{\shortname (DKF)}.
In this setting, only two keyframes will be used to generate the long video based on the text prompt composed by text-based motion planning. 
The results in \cref{fig:video_comp,tab:video_quality_metrics} indicate that although we do not optimize the parameters of the video generative model, our keyframe prompting strategy can significantly preserve the character \& scene identity over a longer video duration.
Please find more details in the video results in supplementary materials.





\vspace{-0.4cm}
\paragraph{Conclusion:}
\shortname is a training-free approach that generates personalized, controllable HSI videos by leveraging 3D-aware keyframe prompting on off-the-shelf video generative models.
Instead of adding new modules to large video generation models with heavy training, our key insight is to divide complex HSI video generation tasks into three stages of script writing, pre-visualization, and animation to generate 3D-aware keyframes and atomic parsed actions for video synthesis.
Results demonstrate the efficacy of our single-view 3D-aware keyframing approach for enhancing 3D human-object contact and subject consistency in HSI video generation.
\vspace{-0.4cm}
\paragraph{Limitations \& Future Work:} 
Our work is limited by the capabilities of existing image inpainting models in inserting humans at a plausible scale and feed-forward Gaussian avatar generation in high-fidelity appearance modeling.
When generating 3D-aware keyframes, we naively render the composed 3D Gaussians of the avatar scene.
Although visual quality could be further enhanced by large image editing models for harmonization and lighting effects, our lower-quality keyframe prompts still yield substantial improvements in HSI video generation.

\noindent\textbf{Acknowledgment}:
This work was supported by NSF CAREER grant \#2143576, a grant from Meta Reality Labs, and an AWS Cloud Credits award.
{
    \small
    \bibliographystyle{ieeenat_fullname}
    \bibliography{main}
}
\clearpage
\appendix

\section{Implementation Details}
All our experiments are conducted on a single Nvidia A6000 GPU. 
Given the user input of image scene and text prompts, we used ChatGPT-4o to conduct all the VLM parsing, including the screenwriting stage and VLM-based CoT contact reasoning.
For each interactive action in the parsing results, we will generate the 3D-aware keyframes.
Before pose generation, we first need to use Trellis~\cite{xiang2024structured} to reconstruct the canonicalized target object, which costs around 1 minute.
Once reconstructed, the object can be reused in all related interaction generation.
We use the Orientate-Anything~\cite{wang2024orient} to canonicalize the object.
Using the canonical view rendering results, the diffusion inpainting model~\cite{inpainting-model} uses 50 steps to inpaint the human on it, which costs around 15 seconds.
For 3D human lifting, we used the AdamW optimizer with $lr=1e^{-3}$ for 1.5k steps, and it only costs less than 2 minutes.
To place the 3D human from object space to scene space, we need to estimate the reconstructed object pose.
Then, we sample the candidate poses based on the initial guess and use the general visual features extracted by DINOv2~\cite{oquab2023dinov2} to measure the similarity between the input object image and the candidate views' observation.
After matching the reconstructed 3D object based on the input object image, we need to transform the object from the object space to world space.
Since the reconstructed 3D object from Trellis has both mesh representation and 3D Gaussian representation, we refine the 6D pose of the object to place it into the scene, aligning it with the original scene image based on RGB constraints via 3D Gaussian rendering and depth constraints via mesh rasterization. 
The ground truth depth of the input image is estimated from MoGe~\cite{Wang2024MoGeUA}.

\section{Non-contact HSI Case}
In some cases, the character is not in contact with the object, such as ``standing next to the chair''; consequently, we will not receive any contact cues from VLM CoT reasoning.
To impose spatial constraints on interactions, we reformulate the interaction loss~$\mathcal{L}_{hoi}$ to constrain the minimal distance $d_{HO}$ between the human mesh and the object mesh, measured in the normalized human mesh space rather than world space:
\begin{equation}
    \mathcal{L}_{hoi}=\begin{cases}
        0, &\quad d_{HO}\leq \delta \\
        d_{HO}-\delta, &\quad d_{HO}>\delta
    \end{cases}
\end{equation}
where $\delta = 10 cm$.

\section{Unnatural Avatar Appearance}
\begin{figure}[ht]
    \vspace{-0.06in}
    \centering
    \includegraphics[height=2.4cm]{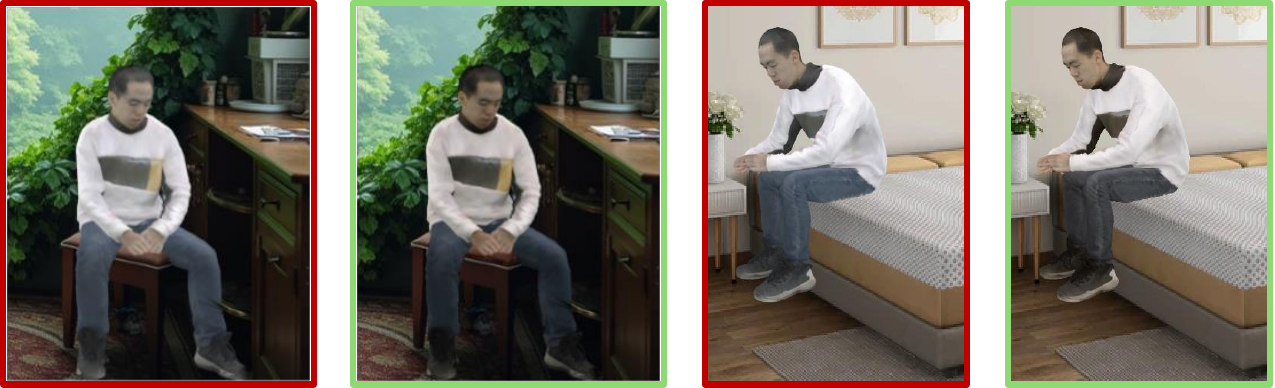}
    \vspace{-0.1in}
    \caption{\footnotesize~Harmonization (\textcolor[rgb]{0.554,0.847,0.449}{green}) significantly improves our appearance.
    \label{fig:harmonization}
    }
    \vspace{-0.3in}
    
\end{figure}

One of the limitations of our work is that the avatar often looks ``plastic'' in the generated video, due to the independent modeling between the 3D Gaussian Avatar and the 3D Gaussian scene.
Although these lighting differences between the scene and the avatar are not the problem we focused on, we now include a harmonization step~\cite{Harmonizer} that significantly improves appearance (\cref{fig:harmonization}).
We also believe that this problem can eliminate artifacts in most existing composition image harmonization methods through post-processing.
Sometimes, the human scale in the generated video is implausible which is the limitation of the image inpainting model. 

\section{Human Body Surface Partition Definition}
To encourage ChatGPT outputs faithful and executable results, we divide the surface of the human body into 15 parts based on SMPL-X~\cite{SMPL-X:2019} template, \ie, ``head'', ``left upper arm'', ``right upper arm'', ``left forearm'', ``right forearm'', ``left hand'', ``right hand'', ``back'', ``buttocks'', ``left thigh'', ``right thigh'', ``left calf'', ``right calf'', ``left foot'', and ``right foot''.

\section{2D Human Insertion Baseline}
We add a 2D baseline based on Flex\footnote{\url{https://huggingface.co/ostris/Flex.2-preview}} with the pose generated from GenHSI.
Flex changes the contents in the scene (\cref{fig:fail2}) and is time-consuming (30s vs 0.1s for 3DGS) to create the keyframes, which is inefficient for long video generation via keyframe interpolation. 
Note that GenHSI only uses 3D to constrain plausible affordances and rendering, and it will not decrease the controllability of pose generation.
Different with feed-forward inpainting solution, our inpainting solution can detect human inpainting results during the denoising loop, effectively serving as an automatic verifier of the inpainting process.
\begin{figure}[ht]
    \centering
    \begin{subfigure}[b]{0.45\linewidth}
         \centering
         \includegraphics[height=2.6cm]{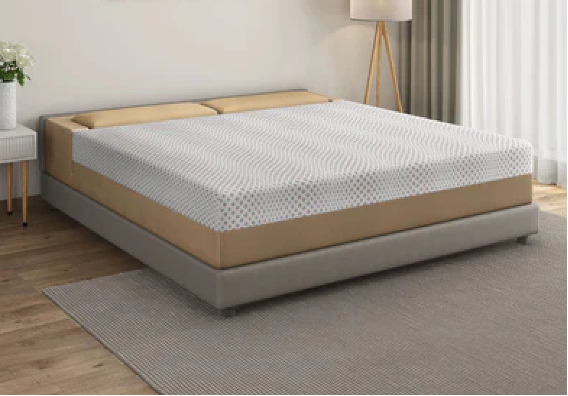}
         \caption{Scene Image}
         \label{fig:fail1}
         \vspace{-0.3cm}
     \end{subfigure}
     \begin{subfigure}[b]{0.45\linewidth}
         \centering
         \includegraphics[height=2.6cm]{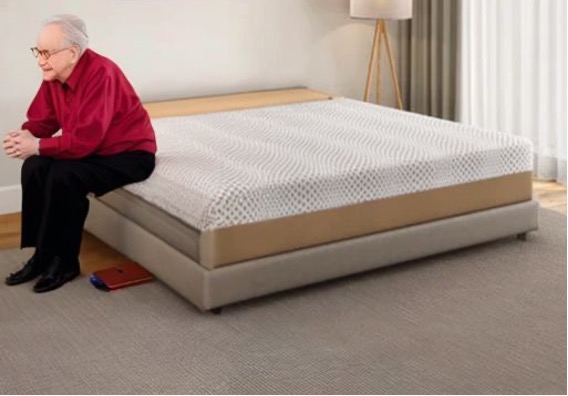}
         \caption{2D HSI Image}
         \vspace{-0.3cm}
         \label{fig:fail2}
     \end{subfigure}
    \label{fig:2D_human_insertion}
\end{figure}

\begin{figure*}
    \centering
    \vspace{-0.6cm}
    \includegraphics[width=\linewidth]{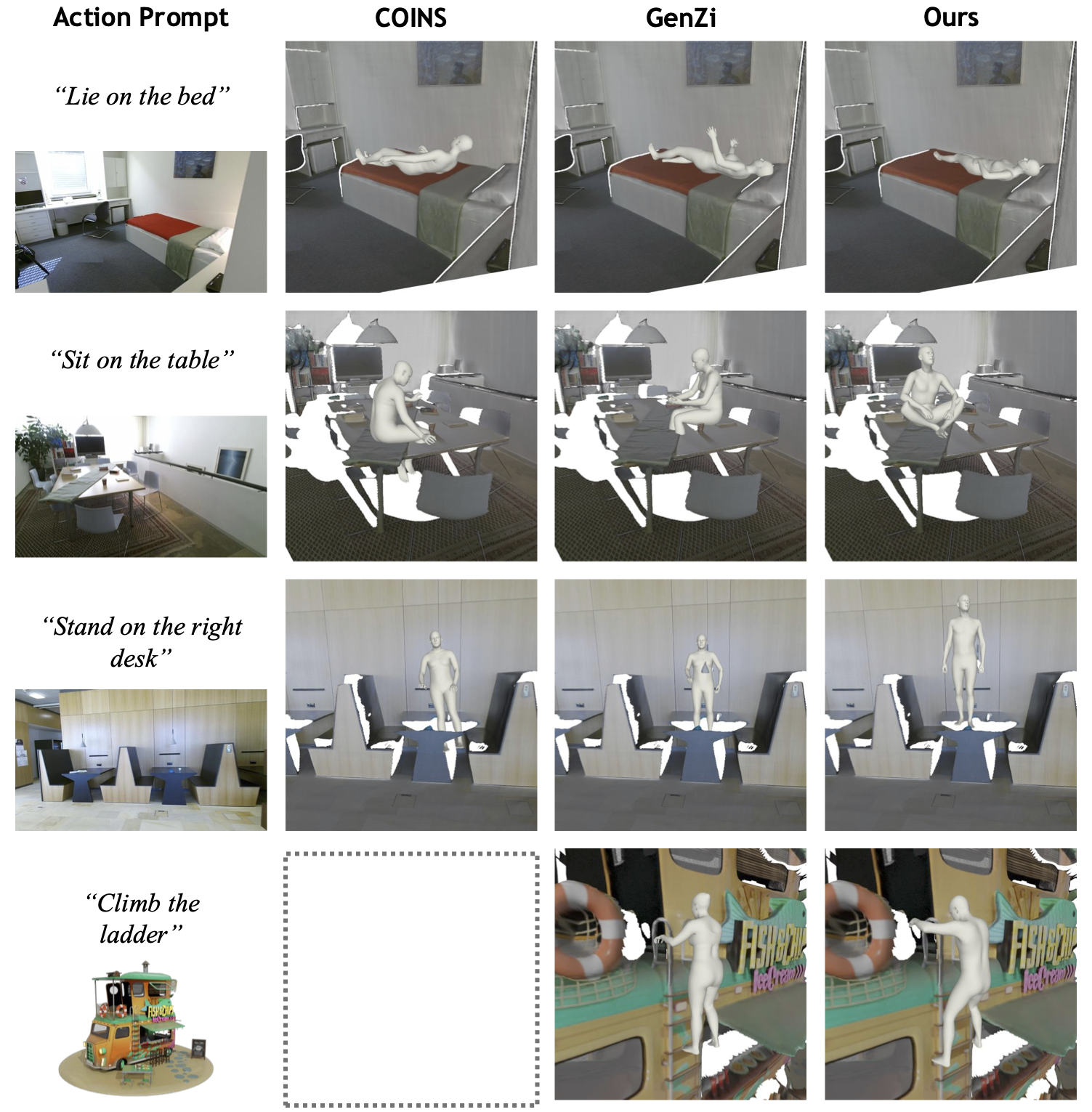}
    \vspace{-0.7cm}
    \caption{\textbf{3D Human-Object Interactions} 
    \shortname performs improved human object interactions even when we don't have access to accurate scene geometry.
    Our work also produces more plausible poses for lying, sitting, and standing. 
    Prior works like GenZI have inconsistent multiview inpainting resulting in diverse but uncomfortable human poses as seen in lying down and sitting on table.
    }
    \label{fig:pose-res}
    \vspace{-0.65cm}
\end{figure*}

\section{Qualitative Results of 3D human Scene Interaction with incomplete 3D Scene}
We also test previous methods in our challenging case, \ie, the single-view image of the scene is the only accessible input about the environment, shown as \cref{fig:pose-res}.
We still use MoGe~\cite{Wang2024MoGeUA} to reconstruct the coarse scene geometry.
COINS~\cite{zhao2022compositional} will use this incomplete scene reconstruction as its point net inputs.
GenZI~\cite{li2024genzi} will render the multi-view image based on the MoGe~\cite{Wang2024MoGeUA} output.
Neither will execute the penetration terms in human joint pose optimization, since they rely on accurate scene geometry.
The results show that our method is more robust based on object-centric human inpainting than GenZI~\cite{li2024genzi}, especially when the scenes are not well structured and well painted.
Also, since we reconstruct the object, even when there is no visible affordance in the input single view, we can still generate plausible human interaction results.

\end{document}